\documentclass[10pt, a4paper]{article}

\usepackage[]{lrec2026} 

\usepackage{makecell}
\usepackage{multirow}
\usepackage{multicol}
\usepackage{arydshln}
\usepackage{amsmath}
\usepackage{cleveref}
\usepackage{subcaption}

\title{Link Prediction for Event Logs in the Process Industry}

\name{Anastasia Zhukova\textsuperscript{1}, Thomas Walton\textsuperscript{2}, Christian E. Lobmüller\textsuperscript{2}, Bela Gipp\textsuperscript{1}} 

\address{\textsuperscript{1}University of Göttingen, Germany, \textsuperscript{2}eschbach GmbH, Germany \\
         \{anastasia.zhukova, gipp\}@uni-goettingen.de, christian.lobmueller@eschbach.com\\}

\abstract{
In the era of graph-based retrieval-augmented generation (RAG), link prediction is a significant preprocessing step for improving the quality of fragmented or incomplete domain-specific data for the graph retrieval. Knowledge management in the process industry uses RAG-based applications to optimize operations, ensure safety, and facilitate continuous improvement by effectively leveraging operational data and past insights. A key challenge in this domain is the fragmented nature of event logs in shift books, where related records are often kept separate, even though they belong to a single event or process. This fragmentation hinders the recommendation of previously implemented solutions to users, which is crucial in the timely problem-solving at live production sites. To address this problem, we develop a record linking model, which we define as a cross-document coreference resolution (CDCR) task. Record linking adapts the task definition of CDCR and combines two state-of-the-art CDCR models with the principles of natural language inference (NLI) and semantic text similarity (STS) to perform link prediction. The evaluation shows that our record linking model outperformed the best versions of our baselines, i.e., NLI and STS, by 28\% (11.43 p) and 27.4\% (11.21 p), respectively. Our work demonstrates that common NLP tasks can be combined and adapted to a domain-specific setting of the German process industry, improving data quality and connectivity in shift logs.
 \\ \newline  }

\begin{document}

\maketitleabstract

\section{Introduction}
In the process industry, knowledge management is a critical component for optimizing operations, ensuring safety, and fostering continuous improvement by capturing, sharing, and utilizing knowledge gained from past experiences \cite{Chua2009}. Knowledge management enables organizations to manage valuable information such as production processes, troubleshooting solutions, and machine performance, which can be used to improve decision-making, reduce errors, and enhance productivity. Retrieval-Augmented Generation (RAG) has emerged as one of the most widely adopted modern architectures for knowledge management applications, combining information retrieval with LLMs to generate responses grounded in retrieved data \cite{Lewis2020}. Many contemporary RAG implementations, e.g., in domain-specific applications, incorporate graph-based retrieval mechanisms that leverage structured knowledge graphs to guide or constrain the retrieval process \cite{barry-etal-2025-graphrag}. Relying on a knowledge graph, especially in low-resource languages, helps LLMs compensate for not knowing a specific domain and terminology well, since they were not trained on the domain's proprietary data.

\begin{figure}
    \centering
    \includegraphics[width=1\linewidth]{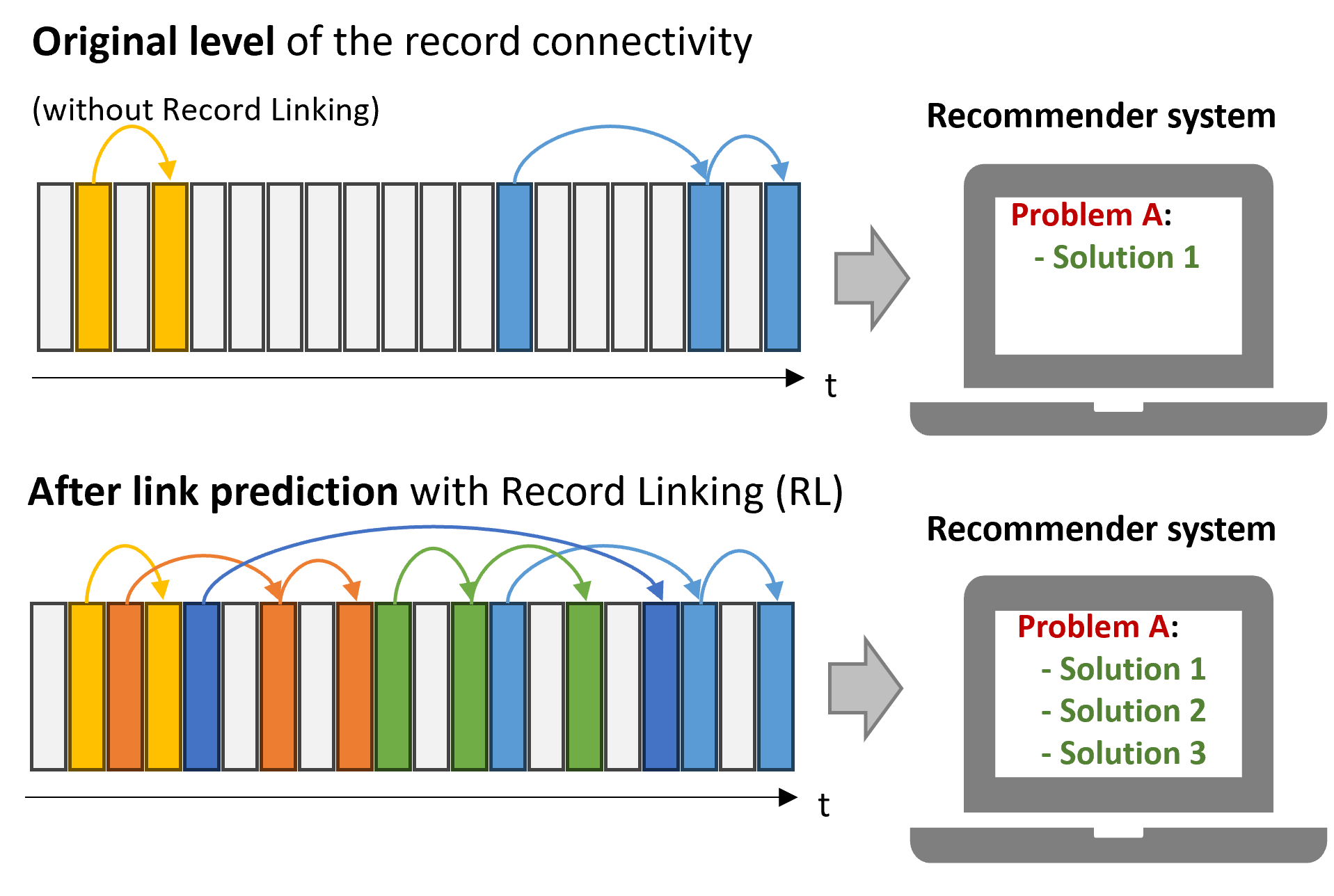}
    \caption{Efficiency and accuracy of the knowledge management applications, such as RAG as a domain-specific solution recommender system, strongly rely on the record connectivity in a knowledge graph. Record linking performs a preprocessing step for link prediction in text logs that report on tasks, problems, and solutions in the production plant, linking records that are part of the same story but were reported as updates to the event. }
    \label{fig:teaser_fig}
\end{figure}

A solution recommender system, as a knowledge management application in the process industry domain, relies on the completeness and connectivity of event logs in the production plant to ensure robust, trustworthy decision support for time-pressing problem-solving tasks (\Cref{fig:teaser_fig}). Text logs of daily operations contain information about tasks, events, and maintenance, as well as previously reported problems and solutions \cite{Zhukova2024}. A problem with the non-linked text logs occurs when two records may be logged as two separate entries, for example, progressively as more details on an event appear, but due to the lack of technical implementation or human factor, remain undocumented via the software interface. Therefore, improving the quality, consistency, and connectivity of the underlying linked data is required to ensure the effective use of the collected domain knowledge in the RAG systems.

\begin{figure*}[h]
    \centering
    \includegraphics[width=1.0\linewidth]{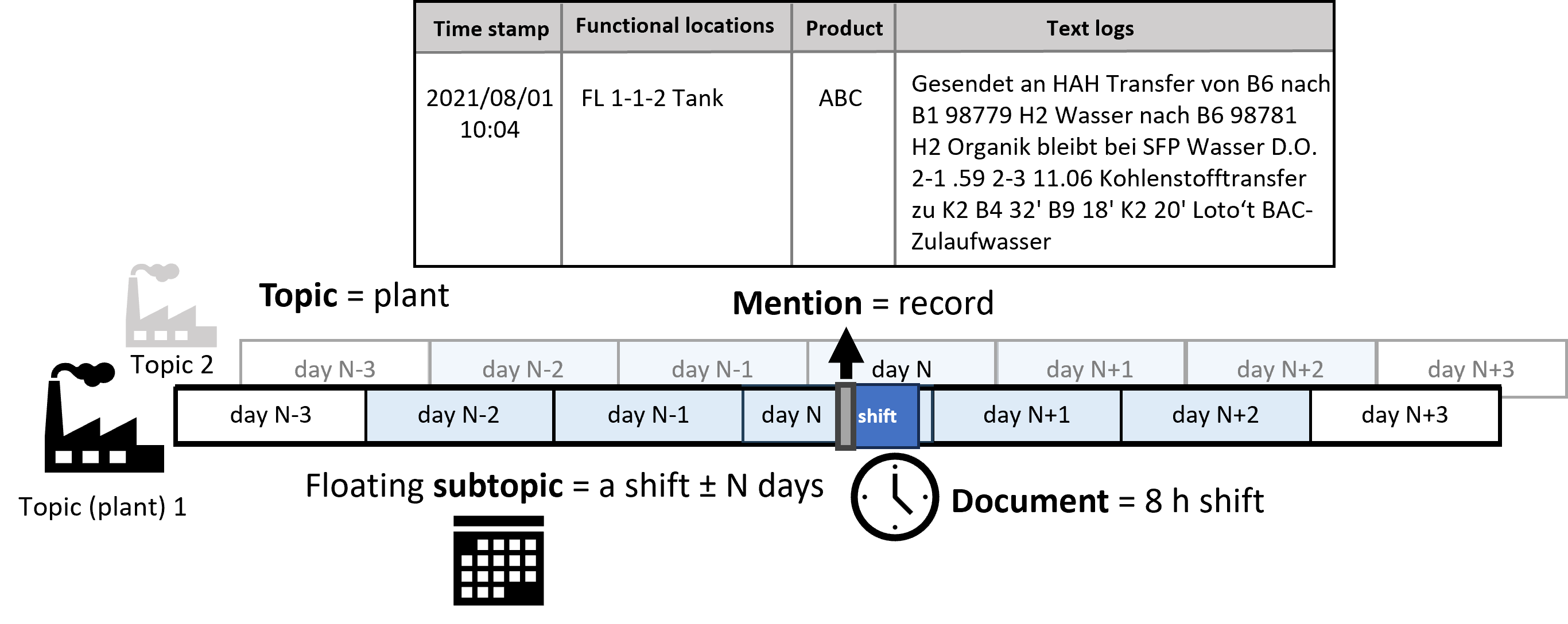}
    \caption{Mapping of CDCR definitions to the record linking task.}
    \label{fig:cdcr_setup}
\end{figure*}

Link prediction in natural language processing (NLP) is commonly referred to as a relation extraction task, in which relationships between entities are identified and extracted from a given text. Although relation extraction is widely applied in tasks such as knowledge graph construction and summarization, relation extraction aims to identify relations between entities, e.g., person-location. To address \textit{the event-driven nature of full-text logs in the process industry}, where multiple logs collectively form a narrative of how an issue is resolved through a series of logically connected events and actions, this paper explores the potential of defining link prediction for record linking as the intersection of several NLP tasks:  event cross-document coreference resolution (CDCR or ECR), natural language inference (NLI), and semantic text similarity (STS). While CDCR focuses on resolving the event-level nature of the logs, NLI and STS address sentence- or passage-level text similarity.

The primary contribution of this paper is to explore and evaluate how common NLP tasks, such as CDCR, NLI, and STS, can be adapted for a specific domain: a link prediction task aimed at improving data quality and connectivity within German logs of daily operations in the process industry. Specifically, we investigate how to combine modifications to CDCR models, adapted for passage-level mentions using NLI and STS, to create a record-linking model. The experiments demonstrate that our CDCR-driven record linking model, built on a domain-adapted GBERT-base, outperforms the best NLI- and STS-driven baselines by 28\% (11.43 points) and 27\% (11.21 points), respectively.

\section{Background}

\subsection{Record linking as NLP tasks}

Link prediction is a task commonly used in graph-based machine learning and network analysis, where the goal is to predict whether a link exists or will form between two nodes. Several NLP tasks focus on identifying relationships and similarities between text spans, including RE \cite{angeli-etal-2015-leveraging}, NLI \cite{bowman-etal-2015-large}, STS \cite{agirre-2012-semeval}, and CDCR \cite{mayfield2009cross}. Relation extraction is the most common NLP task addressing link prediction between entities, such as \textit{"is\_a"} or \textit{"part\_of"}, but it is less common for link prediction between events. NLI is applied in logical reasoning tasks, where it checks whether a claim or answer follows from the given information. STS is commonly used in document similarity assessment tasks to determine the degree of relatedness between two pieces of text.  CDCR is a well-established NLP task that aims to link mentions referring to the same events or entities into chains or clusters of coreferential mentions (e.g., \citet{bugert-gurevych-2021-event, eirew-etal-2021-wec, cattan-etal-2021-cross-document, nath-etal-2024-multimodal, gao-etal-2024-enhancing, chen-etal-2025-employing})\footnote{For example, in the sentences "The President \underline{\textit{announced a new economic policy}} aimed at boosting the national economy" and "\underline{\textit{This initiative}} is expected to create thousands of new jobs across the country," the underlined mentions refer to the same event of the announcement.}. Among these tasks, CDCR not only identifies the strength of the relationship between two text fragments but also groups them into clusters of related elements. Record linking builds on CDCR's methodology, adapting it to handle larger text fragments, such as sentences and passages.

\subsection{Mapping CDCR to record linking}
\label{sec:background}

This section examines how CDCR definitions can be mapped to the record linking task within the process industry domain, as illustrated in \Cref{fig:cdcr_setup}. Record linking aims to identify and link records (or entries) that refer to the same underlying event or process within shift books in the process industry. While CDCR and record linking tasks aim to identify related entities, the record linking task in this domain focuses on larger text fragments, such as sentences or entire passages, rather than phrases, and record linking seeks to link related texts, treating them as parts of a cohesive narrative.

CDCR defines a \textbf{topic} as a shared theme among the documents, e.g., an economic crisis. The topic provides the semantic context that enables the system to associate these mentions, facilitating the resolution of coreferences across documents. In record linking, a topic is represented by a logbook of daily operations from a single production plant. \textbf{Subtopic} represents a specific event within a topic, e.g., the economic crisis of 2008 and the stock market crash of 2025. Subtopics belong to a particular time frame and are often defined by a set of actors, actions, and locations \cite{cybulska-vossen-2014-using}, limiting the document space among which coreferences are to be resolved \cite{barhom-etal-2019-revisiting}. In record linking, we use a subtopic as a sliding window over multiple days, during which issues are typically resolved or directives are addressed.  While a document in CDCR is a news article, we define a \textbf{document} in record linking as an 8-hour production shift. A shift is a defined period that consists of logically connected tasks and events, with a clear beginning and end, much like a structured text document. 

In CDCR, a \textbf{mention} is a phrase in a text that refers to an entity or event, e.g., "Donald Trump" or "the president". In record linking, a mention is \textit{an event record} from a logbook that describes a maintenance event, the current state of production, reports a problem, or provides a solution to it, e.g., as shown in \Cref{fig:cdcr_setup}. Unlike CDCR, where a mention is a word or a phrase, a mention in record linking is a sentence, a paragraph, or a short text and contains structural metadata, such as a timestamp and the code of the machinery it describes. In this work, we will use the terms "mention" and "record" interchangeably. 

In coreference resolution, anaphora defines linguistic expressions that refer to another word or phrase of the same entity or event that form \textbf{coreference relations} \cite{huang2000anaphora}. In turn, mentions in record linking are elements of a single story or issue that are time-structured and logically follow each other; i.e., as NLI defines it, a premise is a previous statement or proposition from which another, a hypothesis, is inferred or follows as a conclusion. In record linking, we define a \textit{coreference relation} as a relation between a premise and a hypothesis. Mentions that belong to one story or incident form a \textbf{coreference chain}. We use a definition of a coreference chain that accounts for the order dependency between mentions, unlike CDCR's mention clusters. A coreference chain can be of the following configurations: (1) \textit{premise (P) - hypothesis (H)}, i.e., a chain of two mentions; (2) \textit{P-H-…-H}, i.e., a chain with multiple mentions that reported follow-ups to a story; (3) \textit{P or H}, i.e., a \textit{singleton} with no follow-up on a story.

\section{Methodology}
Our record linking model adapts and expands upon two CDCR models of \citet{bugert-gurevych-2021-event} and \citet{barhom-etal-2019-revisiting} and consists of the following stages (\Cref{fig:model_architecture}): (1) \textit{a record-pair scoring model} that computes an affinity score for mention pairs, which evaluates the similarity between two potential mentions and determines the likelihood that they refer to the same entity or event, and (2) \textit{mention clustering}, where the previously computed affinity scores are used to group related mentions into clusters, thereby resolving coreference.

\subsection{Record-pair scoring}

Similar to CDCR, record linking relies on joint mention encoding and scoring \cite{lee-etal-2017-end,cattan-etal-2021-cross-document}, where a vector representation of each mention pair is central to the scoring model. The state-of-the-art approach for encoding similarity between two mentions involves combining their contextual information \cite{zeng-etal-2020-event,yu-etal-2022-pairwise,caciularu-etal-2021-cdlm-cross} and enhancing this with additional feature vectors based on mention attributes \cite{barhom-etal-2019-revisiting}. 

\begin{figure}[h]
    \centering
    \includegraphics[width=\linewidth]{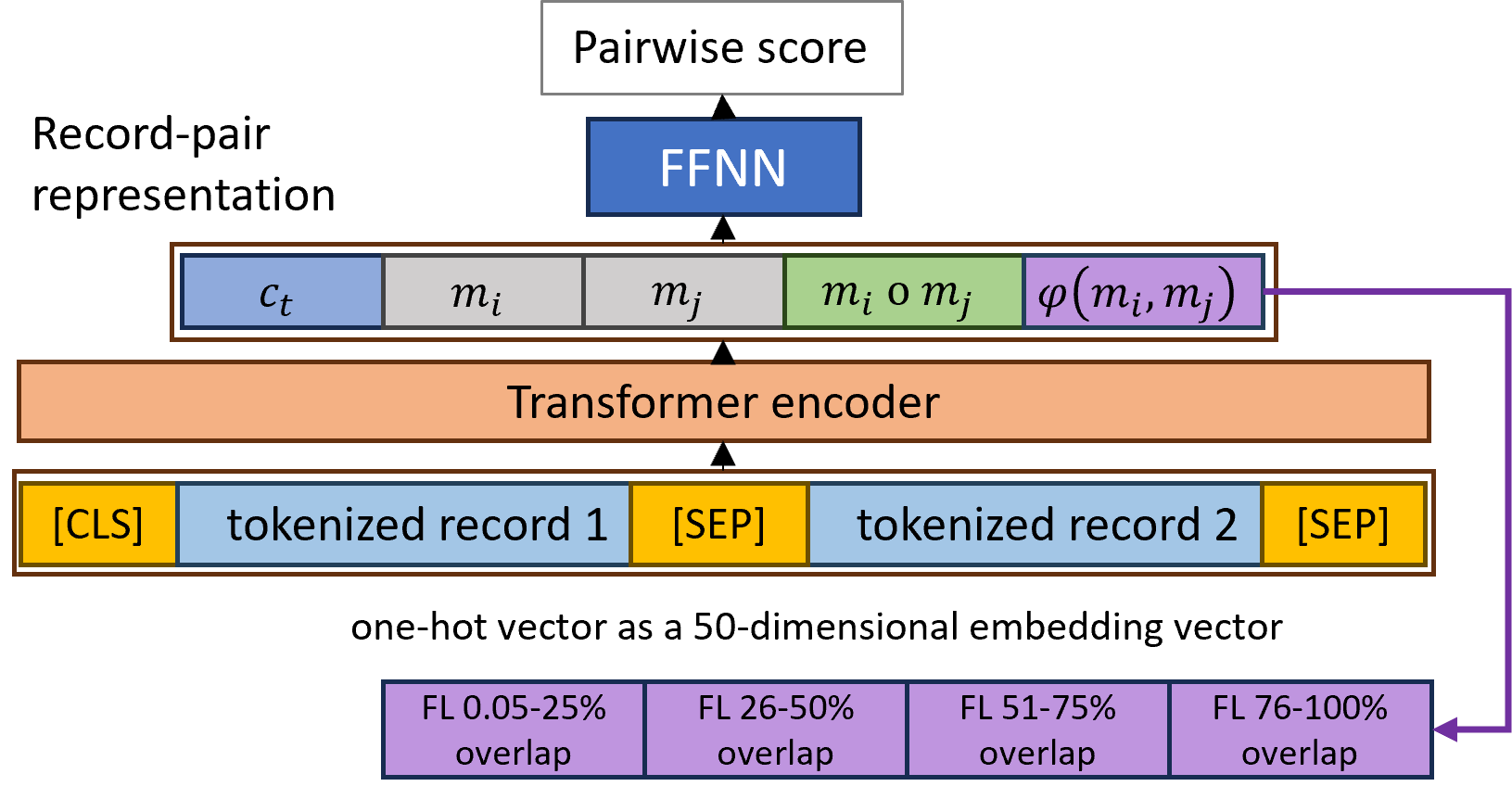}
    \caption{The proposed CDCR-driven record linking model. Compared to most of the state-of-the-art CDCR models \cite{cattan-etal-2021-cross-document, eirew-etal-2021-wec, bugert-gurevych-2021-event}, our joint encoding of the records is enhanced by a joint encoding stemming from the vectors of the [CLS] token \cite{caciularu-etal-2021-cdlm-cross} and a feature vector based on the similarity of the records' attributes \cite{barhom-etal-2019-revisiting}. }
    \label{fig:model_architecture}
\end{figure}

Our record linking method is primarily based on the CDLM model \cite{caciularu-etal-2021-cdlm-cross}, which employs attention-weighted vectors to represent mentions and uses the [CLS] token for jointly encoding concatenated input records. First, two input records are tokenized using a language model’s tokenizer and concatenated into a single sequence formatted as \textit{[CLS] <record 1> [SEP] <record 2> [SEP]}, where the [CLS] token marks the start of the sequence and [SEP] tokens indicate the boundaries between the records. Next, the language model processes this sequence, generating context-dependent vector representations for each token, including the special tokens [CLS] and [SEP]. From these vectors, we extract three key representations: one for the [CLS] token and two attention-weighted pooled vectors corresponding to each mention. Finally, these vectors are combined into a single feature vector 
$m_t(i, j)$, which is fed into a feedforward neural network (FFNN) scorer that outputs a coreference probability or similarity score. The resulting pairwise score is used as a custom affinity metric in clustering to identify coreferential mention chains. \Cref{fig:model_architecture} illustrates our binary classification model, adapted from \cite{cattan-etal-2021-cross-document}, which evaluates the similarity between two mentions. 

The model encodes a mention pair $m_t(i, j)$ as follows:
\begin{equation}
m_t(i, j) = [c_t, m_t^i, m_t^j, m_t^i \circ m_t^j, \varphi(m_t^i,m_t^j)]
\end{equation}
where $c_t$ is a joint mention encoding of two mentions with a transformer model using a CLS token; $m_t^i$ and $m_t^j$ are independent vectors of each mention, which are computed as attention-weighted mean pooling of the corresponding tokens; $m_t^i \circ m_t^j$ is pairwise multiplication of the mentions' vectors; and $\varphi(m_t^i,m_t^j)$ is a feature vector based on the records' attributes that encodes the similarity between functional location (FL) codes, i.e., the codes that refer to the pieces of machinery, about which two mentions report (\Cref{fig:cdcr_setup}). 

Unlike state-of-the-art CDCR models \cite{cattan-etal-2021-cross-document, eirew-etal-2021-wec, caciularu-etal-2021-cdlm-cross, bugert-gurevych-2021-event}, which encode mentions $m_t^i$ and $m_t^j$ as concatenations of the start and end token embeddings, followed by an attention-weighted average, we use only the attention-weighted average, since we use the entire passage rather than a word/phrase as in the original CDCR. This simplification is justified because record linking operates at the passage level, where most mentions typically begin with an article and end with punctuation, making the start and end token embeddings less informative.

The FL feature vector incorporates an external similarity signal based on the overlap between FL codes, in addition to the similarity obtained from the language model. An FL code uniquely identifies a piece of machinery in a production plant and has an agglomerative structure, allowing us to determine if two FL codes share a parent-child relationship or belong to the same family or root. For example, two FL codes \textit{AAAA-CABA-B018} and \textit{AAAA-CABA-A123} share the same parent machinery with a code \textit{AAAA}. The degree of similarity increases with the number of matching characters from the start of the codes, reflecting closer proximity.

We compute the FL similarity as the normalized overlap between two codes:
\begin{equation}
\varphi(m_t^i,m_t^j)=\frac{f_i \cap f_j}{max(len(f_i),len(f_j))}
\end{equation}
This overlap value is then discretized into bins and converted into a one-hot vector corresponding to the assigned bin. To enhance the signal from these binary features, the one-hot vector is passed through an embedding layer with 50 dimensions per bin, following the approach of \cite{barhom-etal-2019-revisiting}.

\subsection{Mention clustering}
The record-linking model employs time-dependent depth-first search (tDFS) mention clustering, which accounts for time constraints between records: two mentions are clustered if they occur within a given time threshold. DFS replaces the state-of-the-art agglomerative clustering (AC) with average linkage \cite{barhom-etal-2019-revisiting, cattan-etal-2021-cross-document, caciularu-etal-2021-cdlm-cross, bugert-gurevych-2021-event}. While agglomerative clustering ignores the order of mentions, tDFS starts with the first mention in the timeline and greedily searches for coreferential mentions to it, then exhausts the search by finding coreferential mentions to the already resolved ones \cite{Zhukova2021}. The time-dependency constraint limits the mention search space to documents and mentions that belong to a single subtopic (\ref{sec:background}), i.e., two mentions separated by a significant time interval cannot belong to the same story.

\begin{table*}[h!]
\centering
\scriptsize
\begin{tabular}{c|c|c|c|l|c}
\hline
ID & Timestamp & Shift & Func. location & \makecell[c]{Description} & Related to \\
\hline
001 & 2026-01-29 10:47:33 & Shift 1 & A013-DR-330 & Temperature spike detected in reactor chamber. & - \\
\hline
002 & 2026-01-29 11:15:22 & Shift 1 & B716-RX-204 & Routine inspection completed, no anomalies detected. & - \\
\hline
003 & 2026-01-29 15:02:10 & Shift 2 &  C118-MX-118 & Lubrication cycle executed successfully. & - \\
\hline
004 & 2026-01-29 18:10:05 & Shift 2 & A013-DR-330 & Cooling valve recalibrated and flow rate increased. & 001 \\
\hline
005 & 2026-01-29 21:25:41 & Shift 2 & B716-FL-501 & Filter replaced as part of scheduled maintenance. & - \\
\hline
006 & 2026-01-29 22:55:12 & Shift 3 & C514-CN-210 & Conveyor belt misalignment causing material spillage. & - \\
\hline
007 & 2026-01-30 01:05:27 & Shift 3 & A013-DR-330 & Additional insulation added to stabilize temperature fluctuations. & 004 \\
\hline
008 & 2026-01-30 05:20:48 & Shift 3  & C514-CN-210 & Belt realigned and tension adjusted. & 006 \\
\hline
009 & 2026-01-30 06:18:59 & Shift 1  & A013-TK-777 & Tank pressure levels within normal operating range. & - \\
\hline
010 & 2026-01-30 10:02:36 & Shift 1 & C514-MX-118 & Mixer calibration verified and logged. & - \\
\hline
\end{tabular}
\caption{A mocked-up example of the data format used in the experiments. Some records are reported without any follow-up information, while others report on the progress of resolving issues that spanned over time. For better illustration, the text data is provided in English. The more true-to-real version of the data record is exemplified in \Cref{fig:cdcr_setup}. }
\label{tab:example_data}
\end{table*}

\subsection{Training}
Our pairwise scorer $\text{sim}(m_i, m_j)$ compares a mention to all other mentions across all documents within a subtopic. 
The adjacent mentions, i.e., the directly neighboring mentions within one chain, are treated as positive examples. 
Unlike CDCR, where the order of mentions is not important, we take the order of the records into account, as they are logically connected into a single story.
Therefore, the negative examples are defined as (1) mentions from two different chains, (2) mentions in the reverse order of a timeline, and (3) non-adjacent mentions (e.g., in a chain A$\rightarrow$B$\rightarrow$C, the mention pair A$\rightarrow$C will be a negative label, and A$\rightarrow$B and B$\rightarrow$C are positive). 
Following \cite{cattan-etal-2021-cross-document, bugert-gurevych-2021-event}, the negative pairs for the training stage are sampled with the proportion 1:20.
The development set for model training contains 1:1 positive and negative samples to ensure the model's F1-score evaluation is not biased by class imbalance.

The overall score is then optimized using binary cross-entropy loss as follows:

\[
L = -\frac{1}{|N|} \sum_{(m_i, m_j) \in N} y \cdot \log(\text{sim}(m_i, m_j))
\]

where \( N \) corresponds to the set of mention-pairs $(m_i, m_j)$, and \( y \in \{0, 1\} \) is the pair label. The FFNN consists of two hidden layers with ReLU activation. 
Similar to state-of-the-art CDCR models, a language model is used solely for mention encoding and is not fine-tuned during training.

The record linking model and its baselines are trained on a single A100 GPU using the AdamW optimizer with a learning rate of 5e{-5}, a weight decay of 0.1, and an epsilon value of 1e{-5}. Training is performed over 5 epochs. Input records were truncated to fit the 512-token limit. During training, positive samples are constructed from adjacent mentions within the same chain, while negative samples include non-adjacent or reversed pairs, i.e., in the reverse-time order. The tDFS clustering method uses the maximum time interval between records, determined by the third quartile (Q3) of the topic-specific time differences between records (\Cref{tab:stats}). We utilize a custom version of the GBERT-base language model, adapted to the process industry domain through continual pretraining \cite{zhukova-2025-efficient}, referred to as daGBERT.

\section{Experiments}
The record linking evaluation follows the CDCR framework by assessing key components, i.e., language model selection, scoring model architecture, and clustering. Evaluation uses standard CDCR metrics and scoring scripts.

\subsection{Dataset}
The data used for training, development, and testing is proprietary and comes from seven German-speaking plants in the chemistry and pharmaceuticals domains. The data originates from the software databases used for daily operations in the process industry and contains manually assigned links by domain system users. \Cref{tab:example_data} illustrates the data format used in the experiments. While the largest portion of the collected data does not include the manually provided information for newer records related to older ones, those that do form the dataset used to develop the record linking model.

\begin{table*}[h]
\scriptsize
\centering
\begin{tabular}{c|r|r|r|r|r|r|r|r|r|r|r|c} 
\hline
\makecell{Topic \\(plant)  } & \multicolumn{5}{c|}{General Stats}                                & \multicolumn{3}{|c}{Full chain (h)} & \multicolumn{3}{|c}{\makecell{Time between \\records (hours)}} & \multicolumn{1}{|c}{\makecell{Train/Dev/Test splits \\(in records/mentions) }}  \\
\hline
    & Records & \makecell{$\Sigma$Chains} & Chains & Singl. & \makecell{Avg.size} & Q1       & Q2       & Q3       & Q1        & Q2         & Q3         &  \\
\hline
A & 87K & 78K & 4K & 73K & 2.96 & 1.3 & 3.7 & 18.9 & 0.5 & 2.0 & 15.8 & 9579 / 1155 / 1174  \\
B & 17K & 17K & 157 & 17K & 2.92 & 10.0 & 56.4 & 463.8 & 0.2 & 30.5 & 213.5 & - / - / 930  \\
C & 25K & 24K & 554 & 23K & 2.56 & 0.0 & 10.1 & 92.5 & 0.0 & 4.7 & 61.4 & 1013 / 1016 / 1041  \\
D & 223K  & 189K & 27K  & 162K & 2.24 & 9.6 & 33.2 & 157.9 & 5.9 & 24.0 & 120.2 & 8341 / 1103 / 1103  \\
E & 59K & 48K & 7K & 41K & 2.40 & 11.3 & 36.6 & 145.4 & 4.9 & 23.1 & 97.2 & 8121 / 1110 / 1079  \\
F & 10K & 9K & 501 & 8K & 2.99 & 5.3 & 53.5 & 213.9 & 0.0 & 18.0 & 110.8 & 956 / 1090 / 905  \\
G & 32K & 28K & 2K & 26K & 2.97 & 10.9 & 114.6 & 341.9 & 2.6 & 44.9 & 162.6 & 8643 / 1253 / 1188  \\
\hline
Total & 454K  & 395K & 42K  & 353K & 2.72 & 6.9 & 44.0 & 204.9 & 2.0 & 21.0 & 111.7 & 36653 / 6727 / 7420 \\
\hline
\end{tabular}
\caption{An overview of the record linking dataset that consists of the data from seven plants, exhibiting significant diversity in factors such as the temporal distance. This variability makes training the record linking model both challenging and robust.}
\label{tab:stats}
\end{table*}

\Cref{tab:stats} illustrates the data collected for the evaluation and the train/dev/test splits used in the experiments. The table shows that the time intervals between the first and last mentions within each chain vary significantly across sources, which, on the one hand, pose challenges for training the record-linking model, but also contribute to a more robust model by exposing it to diverse data patterns. The data split is 0.8/0.1/0.1; if insufficient, the data is used solely for testing (affected one topic). The dataset ensures that the test set contains 200 chains per topic, each composed of the most recent records to closely reflect the data distribution encountered during model inference in deployment. The number of mentions does not always correspond to the number of chains, as it can vary across chains.

Unlike the state-of-the-art CDCR approach, which trains the model on mentions from one topic at a time before moving to the next, we train our model on a mixture of subtopics (see \Cref{fig:topic_subtopic}). This exposes the model to frequently changing mention pairs from different topics throughout training. In state-of-the-art CDCR datasets, which primarily originate from the news domain, the data distribution across topics is more consistent due to standardized document formats, narrative structures, and writing styles. In contrast, production plants do not adhere to a standardized reporting style, resulting in greater variability in their data. Hence, to prevent the catastrophic forgetting of information learned from one plant, we train a model on a shuffled set of subtopics from several sources.

\subsection{Baselines}
\begin{figure}[h!]
    \centering

    \begin{subfigure}{0.45\textwidth}
        \centering
        \includegraphics[width=\linewidth]{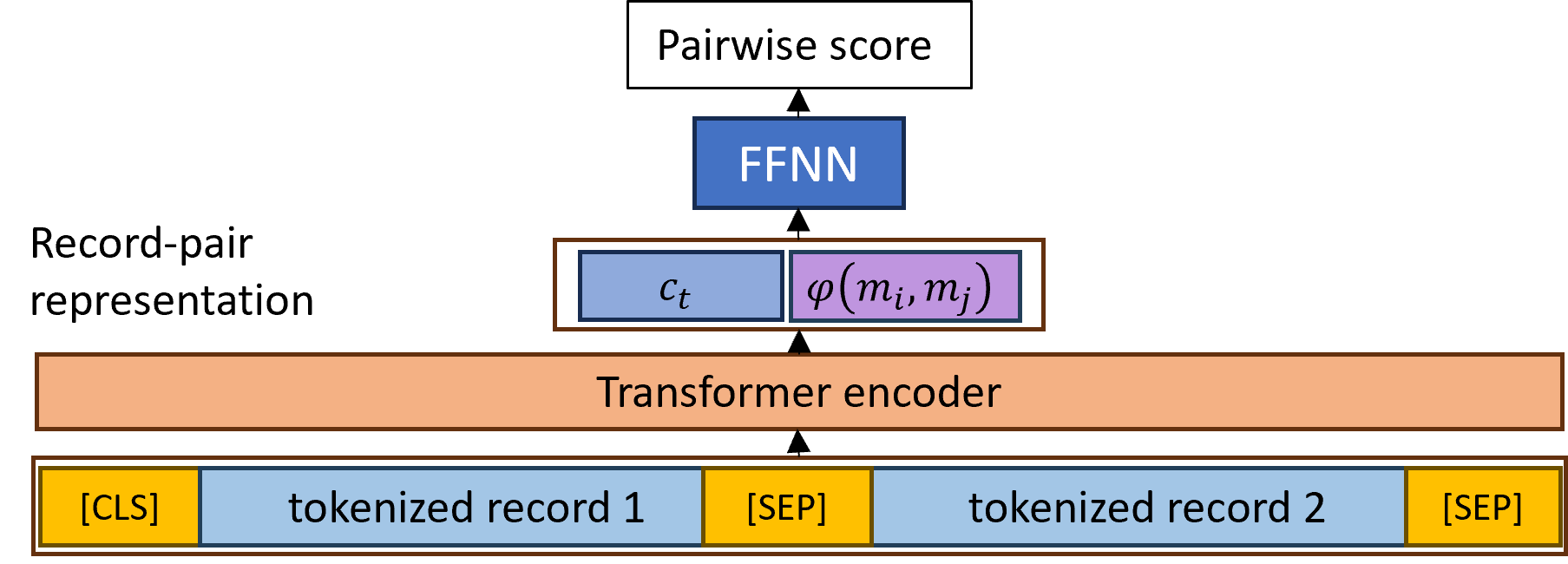}
        \caption{Architecture of the NLI-driven baseline.}
        \label{fig:nli}
    \end{subfigure}
    \hfill
    \begin{subfigure}{0.45\textwidth}
        \centering
        \includegraphics[width=\linewidth]{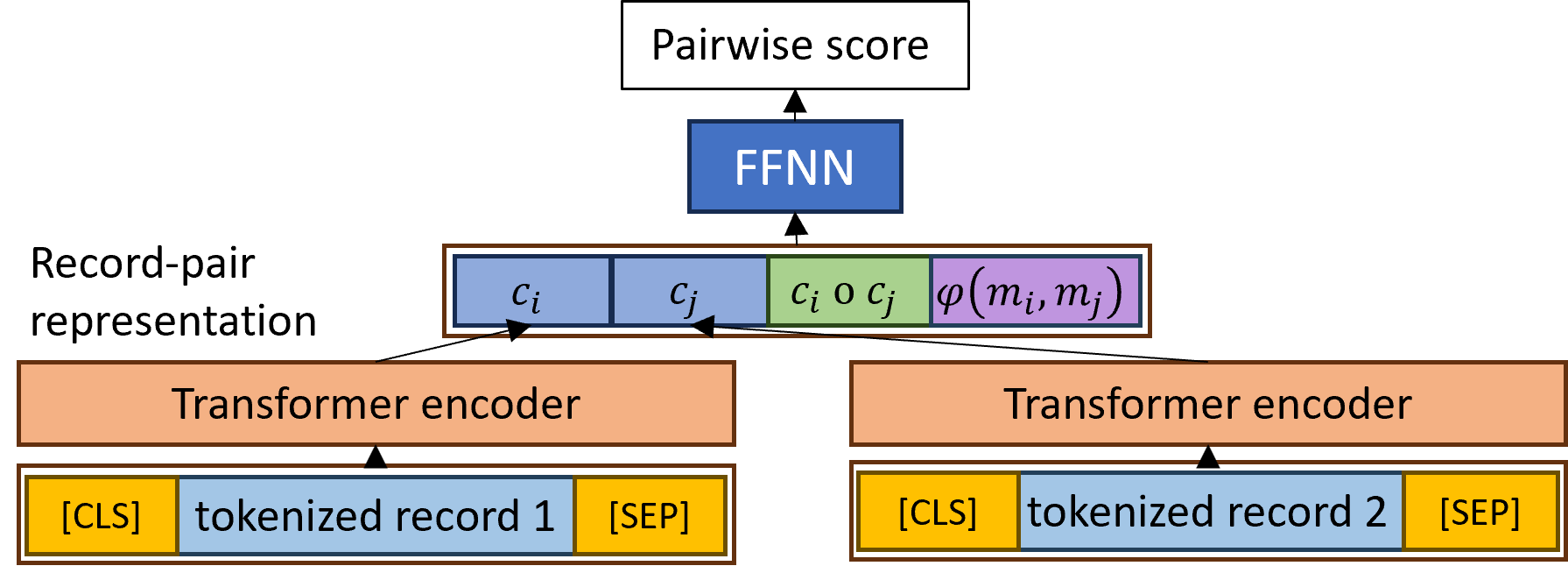}
        \caption{Architecture of the STS-driven baseline.}
        \label{fig:sts}
    \end{subfigure}

    \caption{Baseline architectures. We evaluate the models with and without the feature vector $\varphi$.}
    \label{fig:baselines}
\end{figure}

The baselines are designed to evaluate record linking from three perspectives: (1) the model architectures underlying record linking tasks, specifically NLI and STS (see \Cref{fig:baselines}), (2) the choice of language model used for mention encoding, and (3) the mention clustering method. Additionally, we assess the impact of incorporating the FL feature vector across all model variations.

First, for the NLI-driven architecture (), we employ a joint encoding architecture where two mentions are encoded together using the vector of their preceding [CLS] token \cite{devlin-etal-2019-bert} (\Cref{fig:model_architecture}). In contrast, the STS-driven architecture employs a Siamese network, commonly used in bi-encoder models of sentence transformers \cite{reimers-gurevych-2019-sentence}, which encodes text fragments independently via their [CLS] tokens. These [CLS] vectors serve as mention representations, and pairwise similarity is computed through their element-wise multiplication. 

Second, as baselines for daGBERT, we use two publicly available general-purpose base-sized German language models: (1) the pre-trained \textit{GBERT}-base \cite{chan-etal-2020-germans}, and (2) the best-performing out-of-the-box text encoder selected based on a domain-specific benchmark for semantic search \cite{zhukova-etal-2025-automated}, namely \textit{mGTE} \cite{zhang-etal-2024-mgte}\footnote{\url{https://huggingface.co/Alibaba-NLP/gte-multilingual-base}}. We use mGTE exclusively for the STS-driven architecture, while for GBERT in the STS architecture, we apply mean pooling over the last layer’s hidden states.

Finally, we compare \textit{tDFS} with the agglomerative clustering (\textit{AC}) method commonly used in CDCR. We apply agglomerative clustering using single linkage to align with the requirements for consecutive clustering of mentions into chains, also known as the friends-of-friends algorithm.

\subsection{Evaluation}
\label{sec:metrics}
The record-linking model is an end-to-end pipeline comprising two steps: a pairwise scorer and a clustering step. Therefore, each step in the pipeline needs to be evaluated and optimized separately before evaluating the complete end-to-end pipeline. 

The scoring model is evaluated on the development set using the F1-score for binary classification. First, we select the best model across the epochs based on the lowest loss computed on the development set. The threshold that yields the highest F1-score on the development set, determined via ROC analysis, is selected as the preliminary clustering threshold. Additionally, to assess the overall performance of the scoring models, we computed the F1-score at a cut-off of 0.05. 

For clustering, the threshold is optimized on the development set within a $\pm$30\% to $\pm$100\% range of the selected value. Clustering performance is evaluated using homogeneity, completeness, and v-measure, and the threshold yielding the highest v-measure is selected for testing.

Finally, the end-to-end evaluation of the record linking model is performed on the test set using the best-scoring checkpoint of the scoring model and the optimal clustering threshold determined by CDCR metrics. We use the standard metrics for coreference resolution \cite{pradhan-etal-2012-conll}, such as MUC \cite{10.3115/1072399.1072405}, $B^3$ \cite{Bagga98algorithmsfor}, and CEAF\textsubscript{e} scores \cite{luo-2005-coreference}, and report the average of these scores called the F1 CoNLL score to provide a more comprehensive measure of a system’s real-world performance. We follow the principle of \cite{cattan-etal-2021-realistic} and test CDCR models on test sets without singletons, as both $B^3$ and CEAF\_e metrics have been criticized for inflating scores by giving undue credit to singleton mentions \cite{10.1017/S135132491000029X}.

\begin{table*}[]
\scriptsize
\centering
\begin{tabular}{l|c|c|c|c|ccc|ccc|ccc|c}
\hline
\multirow{2}{*}{} & \multirow{2}{*}{LM} & \multirow{2}{*}{FL} & \multirow{2}{*}{\makecell{F1- \\ score}} & \multirow{2}{*}{Clust.} & \multicolumn{3}{c|}{MUC} & \multicolumn{3}{c|}{$B^3$}  & \multicolumn{3}{c|}{CEAF\textsubscript{e}}  & \multirow{2}{*}{\makecell{F1 \\ CoNLL}} \\
 &  &  & & & R  & P  & F1  & R  & P  & F1  & R  & P  & F1  &  \\
 \hline
 &  & & & AC & 100.00 & 63.08 & 76.91 & 100.00 & 17.01 & 23.91 & 10.47 & 29.04 & 12.74 & 37.85 \\
 &  & \multirow{-2}{*}{-} & \multirow{-2}{*}{78.83} & tDFS & 0.00 & 0.00 & 0.00 & 42.27 & 100.00 & 59.17 & 60.89 & 26.04 & 36.36 & 31.84 \\
 \cdashline{3-15}
 &  & &  & AC & 100.00 & 63.08 & 76.91 & 100.00 & 17.01 & 23.91 & 10.47 & 29.04 & 12.74 & 37.85 \\
 & \multirow{-4}{*}{GBERT} & \multirow{-2}{*}{+} & \multirow{-2}{*}{76.58} & tDFS & 20.22 & 24.95 & 21.07 & 52.52 & 63.71 & 53.86 & 50.87 & 38.00 & 41.84 & 38.93 \\
 \cline{2-15}
 &  & &  & AC & 100.00 & 63.08 & 76.91 & 100.00 & 17.01 & 23.91 & 10.47 & 29.04 & 12.74 & 37.85 \\
 &  & \multirow{-2}{*}{-} & \multirow{-2}{*}{72.64} & tDFS & 11.45 & 32.29 & 15.84 & 47.74 & 87.78 & 60.99 & 62.01 & 32.30 & 42.16 & 39.66 \\
 \cdashline{3-15}
 &  & & & AC & 100.00 & 63.08 & 76.91 & 100.00 & 17.01 & 23.91 & 10.47 & 29.04 & 12.74 & 37.85 \\
\multirow{-8}{*}{\rotatebox{90}{NLI-driven}} & \multirow{-4}{*}{daGBERT} & \multirow{-2}{*}{+} & \multirow{-2}{*}{68.52} & tDFS & 15.88 & 30.66 & 19.69 & 50.07 & 78.13 & 59.34 & 58.66 & 35.07 & 43.25 & 40.76 \\
 \hline
 &  & &  & AC & 99.71 & 62.99 & 76.76 & 99.81 & 17.16 & 24.17 & 10.67 & 32.47 & 13.12 & 38.02 \\
 &  & \multirow{-2}{*}{-} & \multirow{-2}{*}{67.34} & tDFS & 25.31 & 28.33 & 25.62 & 55.34 & 61.81 & 53.26 & 49.57 & 39.28 & 41.77 & 40.22 \\
 \cdashline{3-15}
 &  & & & AC & 99.90 & 63.04 & 76.85 & 99.93 & 17.06 & 23.99 & 10.53 & 30.59 & 12.87 & 37.90 \\
 & \multirow{-4}{*}{GBERT} & \multirow{-2}{*}{+} & \multirow{-2}{*}{66.64} & tDFS & 27.88 & 29.11 & 27.46 & 56.96 & 55.03 & 51.41 & 46.69 & 41.63 & 41.65 & 40.17 \\
 \cline{2-15}
 &  &  & & AC & 100.00 & 63.08 & 76.91 & 100.00 & 17.01 & 23.91 & 10.47 & 29.04 & 12.74 & 37.85 \\
 &  & \multirow{-2}{*}{-} & \multirow{-2}{*}{66.46} & tDFS & 10.55 & 20.37 & 13.05 & 47.52 & 78.97 & 57.97 & 58.95 & 33.78 & 42.40 & 37.81 \\
 \cdashline{3-15}
 &  & & & AC & 100.00 & 63.08 & 76.91 & 100.00 & 17.01 & 23.91 & 10.47 & 29.04 & 12.74 & 37.85 \\
 & \multirow{-4}{*}{mGTE} & \multirow{-2}{*}{+} & \multirow{-2}{*}{65.37} & tDFS & 5.87 & 10.82 & 6.67 & 45.20 & 87.82 & 57.66 & 58.34 & 29.81 & 38.58 & 34.30 \\
 \cline{2-15}
 &  & &  & AC & 99.52 & 62.96 & 76.68 & 99.70 & 17.29 & 24.43 & 10.89 & 34.35 & 13.53 & 38.21 \\
 &  & \multirow{-2}{*}{-} & \multirow{-2}{*}{72.52} & tDFS & 24.16 & 30.62 & 25.73 & 54.57 & 62.79 & 54.51 & 50.89 & 39.70 & 42.70 & 40.98 \\
 \cdashline{3-15}
 &  & & & AC & 98.71 & 62.68 & 76.23 & 99.14 & 17.64 & 25.04 & 11.27 & 36.23 & 14.19 & 38.49 \\
\multirow{-12}{*}{\rotatebox{90}{STS-driven}} & \multirow{-4}{*}{daGBERT} & \multirow{-2}{*}{+} & \multirow{-2}{*}{68.65} & tDFS & 28.20 & 29.38 & 27.60 & 57.06 & 55.91 & 50.96 & 46.32 & 40.39 & 40.81 & 39.79 \\
 \hline
 &  & &  & AC & 99.47 & 62.95 & 76.65 & 99.64 & 17.32 & 24.47 & 10.92 & 34.84 & 13.58 & 38.23 \\
 &  & \multirow{-2}{*}{-} & \multirow{-2}{*}{78.83} & tDFS & 17.98 & 37.02 & 23.03 & 50.92 & 83.56 & 62.23 & 63.40 & 36.98 & 46.11 & 43.79 \\
 \cdashline{3-15}
 &  & & & AC & 97.80 & 62.49 & 75.82 & 98.48 & 18.30 & 26.14 & 12.17 & 40.02 & 15.63 & 39.20 \\
 & \multirow{-4}{*}{GBERT} & \multirow{-2}{*}{+} & \multirow{-2}{*}{78.10} & tDFS & 33.82 & 41.92 & 36.29 & 59.26 & 65.76 & 59.90 & 57.24 & 47.20 & 50.31 & 48.83 \\
 \cline{2-15}
 &  &  & & AC & 92.23 & 60.90 & 72.90 & 94.88 & 21.76 & 31.13 & 15.79 & 43.20 & 20.29 & 41.44 \\
 &  & \multirow{-2}{*}{-} & \multirow{-2}{*}{\textbf{81.05}} & tDFS & 52.70 & 51.75 & 50.14 & 70.65 & 52.76 & 54.06 & 47.48 & 53.09 & 46.77 & \textit{\underline{50.32}} \\
 \cdashline{3-15}
 &  & &  & AC & 92.35 & 60.94 & 72.95 & 95.15 & 21.65 & 30.79 & 15.91 & 43.58 & 20.23 & 41.33 \\
\multirow{-8}{*}{\rotatebox{90}{RL (CDCR-driven)}} & \multirow{-4}{*}{daGBERT} & \multirow{-2}{*}{+} & \multirow{-2}{*}{\textit{\underline{80.51}}} & tDFS & 46.05 & 49.57 & 46.36 & 66.77 & 59.51 & 59.37 & 53.89 & 51.40 & 50.84 & \textbf{52.19 }\\
 \hline
\end{tabular}
\caption{Evaluation results demonstrate that our proposed record linking (CDCR-driven) model consistently outperforms all baseline variants. LM stands for language model. Specifically, the combination of the joint encoder architecture at the mention level, daGBERT for text vectorization, the FL feature vector, and the custom tDFS clustering algorithm achieves the highest performance.}
\label{tab:e2e_results}
\end{table*}

\begin{figure}[h!]
    \centering
    \includegraphics[width=\linewidth]{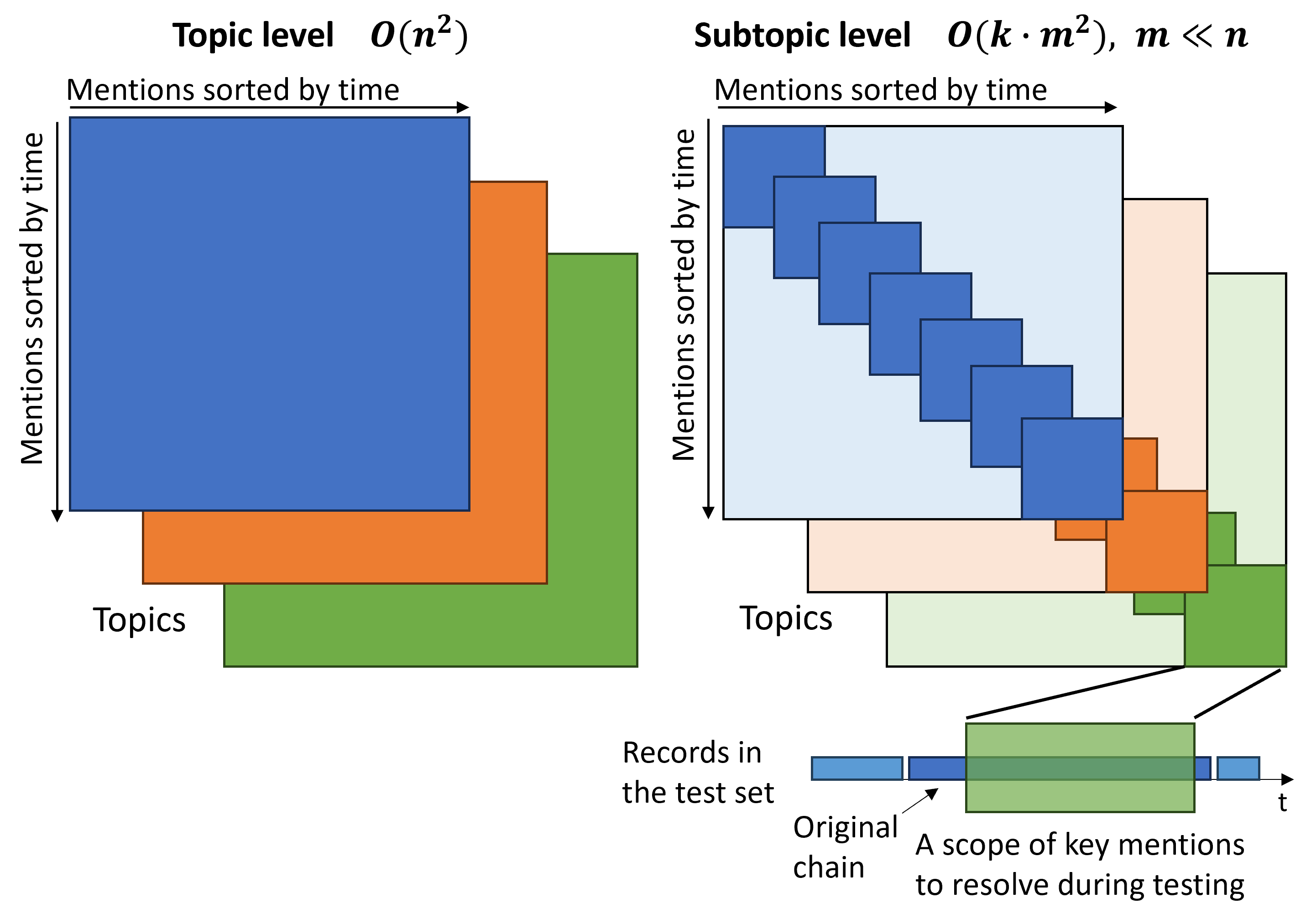}
    \caption{The comparison of evaluation on the topic vs. subtopic level in computational effort in computing the similarity matrices. The subtopic level saves computational effort by avoiding the computation of unnecessary scores between temporally distant mentions. 
    Some original chains may be split by the subtopic time frame; therefore, a sliding subtopic window is required to evaluate all parts of the original chains.  }
    \label{fig:topic_subtopic}
\end{figure}

Evaluation is conducted at the subtopic level (\Cref{fig:topic_subtopic}), followed by aggregation to the topic level through averaging the subtopic results. The subtopic level ensures more efficient computation time compared to the topic level (topic level $O(n^2)$ vs. subtopic level $O(k \cdot m^2), \quad m \ll n$). The window sizes for subtopics are determined based on the time interval of the full chain of each topic (\Cref{tab:stats}). Although the subtopic time frame is defined using the third quartile (Q3) of the full chain time distances per topic (see \Cref{tab:stats}), this approach can result in some chains being split (\Cref{fig:topic_subtopic}). To address this issue, we employ overlapping sliding windows, i.e., overlapping subtopics, to evaluate all parts of the chains, ensuring a comprehensive assessment of how the model resolves mentions across potentially split chains.

\subsection{Results}

\Cref{tab:e2e_results} shows that the proposed record linking model, as a CDCR-driven architecture using daGBERT as the text encoder and tDFS as the clustering algorithm, substantially outperforms the strongest NLI and STS baselines. In particular, it achieves improvements of 28\% (11.43 p) over the best NLI-based variant and 27.4\% (11.21 p) over the best STS-based variant in terms of $F1_{\text{CoNLL}}$. Since the record-linking model comprises several components in the end-to-end pipeline, i.e., CDCR-driven architecture, daGBERT, FL feature vector, and tDFS clustering, we analyze and discuss the effects of each on the final result. 

When comparing the influence of the model architecture, we will use the GBERT as text encoder and agglomerative clustering versions and will look at two metrics: (1) F1-score, i.e., the trained pairwise model evaluated on the binary classification task using the development set, (2) the final $F1_{\text{CoNLL}}$ on the test set. We see that the CDCR-driven architecture outperforms only the STS-baseline and performs as well as the NLI-baseline when using F1-score, while it outperforms 
the baselines with a marginal gain $F1_{\text{CoNLL}}$, i.e., 38.23 vs. 37.85 for NLI and 38.02 for STS. The results suggest that the architectures and training processes do not capture semantic relatedness between records sufficiently when they are not augmented with domain-specific information, such as the FL feature vector and a domain-adapted language model.

We see performance improvements across STS- and CDCR-driven models that use daGBERT, a domain-adapted German language model for the process industry trained via continual pretraining and fine-tuned as a text encoder, and this pattern persists despite using the FL feature vector. Moreover, although mGTE outperformed daGBERT in the semantic search task (see \cite{zhukova-2025-efficient}), it performs the worst among all baselines and modifications when applied to encode records for the record linking task, while daGBERT shows systematic improvement compared to GBERT, measured by both F1-score and $F1_{\text{CoNLL}}$. 

An FL feature vector has a delayed effect on the end-to-end pipeline performance, and only for NLI- and CDCR-driven models paired with tDFS and daGBERT. Specifically, the F1-score decreases when the FL feature vector is used for model training across all baselines and model variations. But the effect becomes pronounced when combined with tDFS, which increases $F1_{\text{CoNLL}}$ in NLI- and CDCR-driven architectures by 5.04-7.09 p, and especially when additionally combined with daGBERT by 8.4-8.92 p. 

Lastly, tDFS systematically outperforms agglomerative clustering in all architectures when the models use GBERT and daGBERT as text encoders. The largest effect of tDFS is seen in the CDCR-driven architecture, where performance improved by 14.5-25.8\% compared to NLI and STS with only a 2.9-7.6\% increase. We see the time constraint as the main factor in the systematic performance increase, since, given the problem definition and data statistics (see \Cref{tab:stats}), the related records cannot be far apart in time, and tDFS accounts for the time difference between them. 

\begin{figure}[h]
    \centering
    \includegraphics[width=\linewidth]{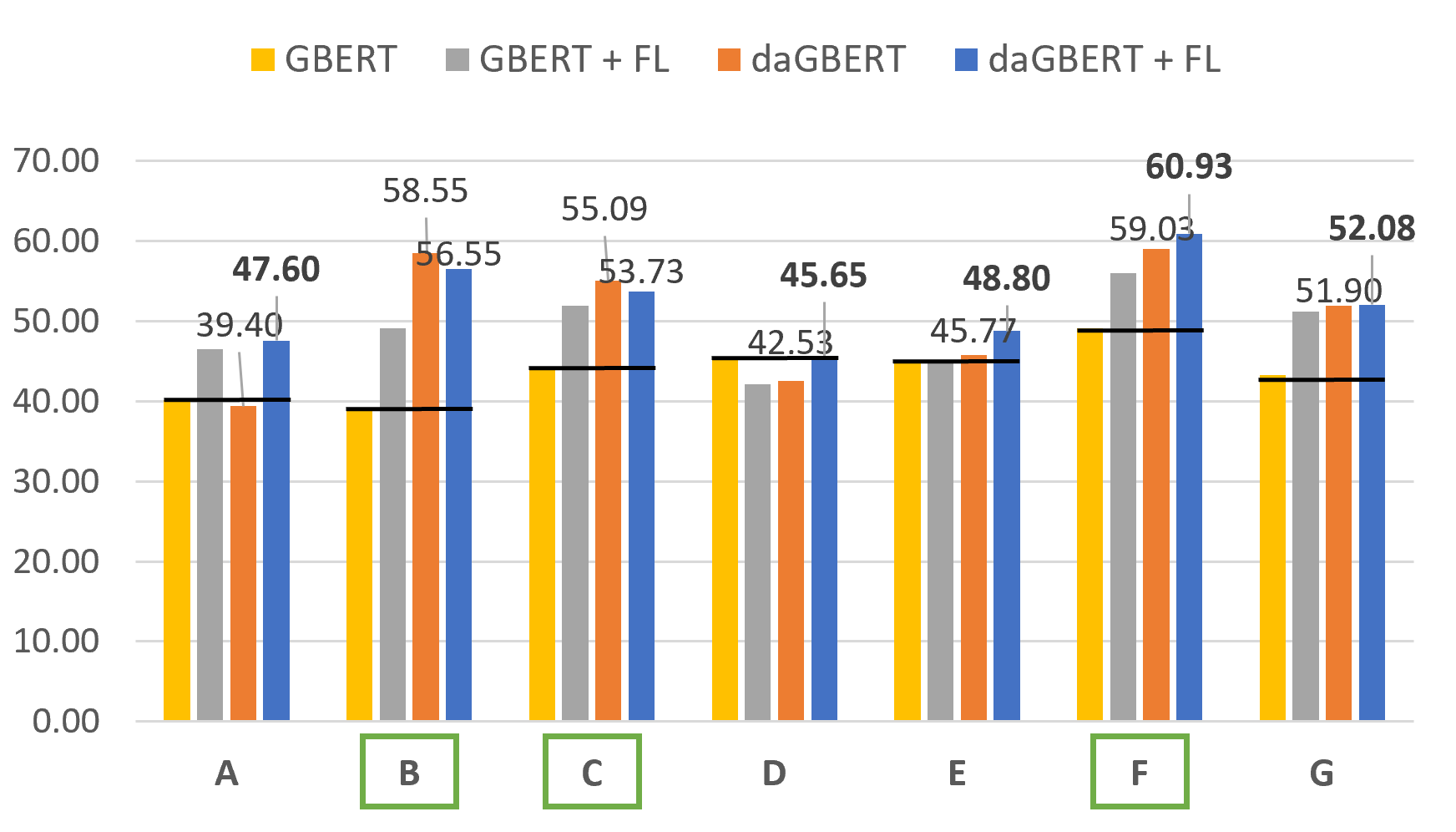}
    \caption{Record linking performance on a topic level. The proposed record linking with daGBERT+FL outperformed all other modifications across almost all topics when using tDFS. }
    \label{fig:rl_topic}
\end{figure}

\Cref{fig:rl_topic} shows the $F1_{\text{CoNLL}}$ scores of all record linking model variants across different topics, highlighting that the daGBERT+FL+tDFS combination outperforms other baselines in 5 out of 7 topics. Additionally, daGBERT consistently outperforms GBERT across nearly all cases, and incorporating the FL feature further improves record linking performance across most topics. The results also demonstrate transfer learning across topics: the record linking model achieves strong performance on Topics B, C, and F, which were either not observed or observed only to a limited extent during training due to limited data, ranking first and second among the other topics.

The impact of FL features is consistently positive but varies in magnitude across topics. For GBERT, adding FL leads to noticeable improvements in more topics than for daGBERT. This indicates that the FL feature vector provides complementary information independent of the encoder, but its contribution depends on how well the base model already captures domain semantics. 

Finally, the results highlight topic-dependent difficulty and variability. Topic D remains the most challenging across all configurations, with comparatively lower performance. Importantly, the relative ranking of models remains stable across topics: daGBERT + FL > daGBERT > GBERT + FL > GBERT, reinforcing the conclusion that both domain adaptation and feature augmentation are robust contributors to performance, regardless of topic variation.

\section{Discussion}

This study demonstrates that CDCR methods can be successfully adapted for record linking within the process industry shift logs, where events and equipment-related issues are described in complex, domain-specific language. A record-linking model serves as a preprocessing step to reconstruct missing links between related text records, which are parts of the same event that were reported with more detail on a given issue. 

The results highlight that improvements in end-to-end record linking performance cannot be attributed to a single component, but rather emerge from the interaction between architecture, representation, and clustering strategy. While the CDCR-driven architecture alone provides only marginal gains over NLI and STS baselines in terms of $F1_{\text{CoNLL}}$, its true advantage becomes evident when combined with domain-specific enhancements. In particular, the discrepancy between pairwise F1-scores and end-to-end CoNLL performance suggests that optimizing local pairwise decisions is insufficient for high-quality clustering. This indicates that architectural differences primarily influence how effectively downstream components, such as clustering, can exploit learned representations, rather than directly improving pairwise classification.

A central finding is the importance of domain adaptation at both the representation and feature levels. The consistent improvements achieved by daGBERT over GBERT demonstrate that task-specific language adaptation is critical for record linking. Similarly, the FL feature vector does not directly improve pairwise classification performance, as it often degrades it, but yields substantial gains in $F1_{\text{CoNLL}}$ when combined with tDFS. This delayed effect suggests that FL features encode global or structural signals that are not immediately useful for binary decisions but become valuable when integrated into clustering, reinforcing the idea that end-to-end performance depends on the combination of feature-algorithm compatibility rather than isolated component quality. Moreover, the usability of the FL feature vector shows that the structured metadata plays an important complementary role as a source of domain-specific information. Future work will focus on enhancing the robustness of the proposed model and on exploring less discrete approaches to feature vectors that encompass a broader range of structural record attributes.

Finally, the results emphasize the pivotal role of clustering, particularly when incorporating temporal constraints. The substantially larger gains observed in the CDCR-driven architecture suggest that it better aligns with the assumptions embedded in tDFS, enabling more effective exploitation of time-aware constraints. Moreover, the strong performance across unseen or low-resource topics points to robust generalization and transfer capabilities, likely driven by the combination of domain-adapted representations and structurally informed clustering. Overall, the findings underscore that effective record linking requires a holistic design, in which architecture, features, and clustering are jointly optimized to reflect the data's underlying characteristics.

Despite the era of LLMs, our methodological choice to rely on BERT-based models is primarily motivated by their comparatively lightweight fine-tuning and inference costs. Base-sized transformer encoders can be efficiently adapted to domain-specific data under limited computational resources, requiring substantially less memory and training time than LLMs. Moreover, at inference time, BERT-based architectures enable faster pairwise scoring, which is critical for scenarios where link prediction must be performed continuously during the indexing of incoming text records. In future work, we will investigate how LLMs can improve record linking while keeping our primary focus on solutions that ensure fast, stable, and cost-efficient inference in low-resource production settings.

\section{Conclusion}

This work demonstrates that record linking is a critical enabler for improving knowledge graph connectivity and, consequently, the effectiveness of knowledge management systems in the process industry. By formulating link prediction as a combination of CDCR, NLI, and STS, we show that capturing the event-driven and temporal structure of industrial logs requires more than sentence-level similarity. The proposed CDCR-driven record-linking model, leveraging a domain-adapted language model (daGBERT), functional-location feature augmentation, and time-aware clustering (tDFS), achieves substantial improvements over NLI- and STS-based baselines, highlighting the importance of jointly optimizing representations, features, and clustering. The results additionally reveal that domain adaptation and temporal constraints are key factors for robust performance, transfer learning, and generalization across topics. Ultimately, enhancing record connectivity strengthens the underlying knowledge graph, enabling more accurate and reliable retrieval in downstream applications, such as an RAG-based solution recommender system that supports better decision-making in time-critical industrial environments.

\section*{Limitations}

This study does not aim to provide an exhaustive exploration of existing link prediction methods and models commonly used in knowledge graph construction. Instead, it focuses specifically on the event-driven nature of industrial production records and investigates how their structural and temporal characteristics align more closely with CDCR tasks. Consequently, the findings primarily reflect the applicability of CDCR-inspired techniques to this particular domain and data type. The proposed approach may not generalize directly to other domains where events, textual structures, or relational patterns differ substantially. Furthermore, the model’s performance may vary depending on factors such as the chosen language model, the set of metadata attributes employed, or the nature and quality of the available records.

\section*{Acknowledgments}
This project is supported by the Federal Ministry for Economic Affairs and Climate Action (BMWK) on the basis of a decision by the German Bundestag.

\section{Bibliographical References}\label{sec:reference}

\bibliographystyle{lrec2026-natbib}
\bibliography{lrec2026-example}

\begin{thebibliography}{34}
\expandafter\ifx\csname natexlab\endcsname\relax\def\natexlab#1{#1}\fi

\bibitem[{Agirre et~al.(2012)Agirre, Diab, Cer, and Gonzalez-Agirre}]{agirre-2012-semeval}
Eneko Agirre, Mona Diab, Daniel Cer, and Aitor Gonzalez-Agirre. 2012.
\newblock Semeval-2012 task 6: a pilot on semantic textual similarity.
\newblock In \emph{Proceedings of the First Joint Conference on Lexical and Computational Semantics - Volume 1: Proceedings of the Main Conference and the Shared Task, and Volume 2: Proceedings of the Sixth International Workshop on Semantic Evaluation}, SemEval '12, page 385–393, USA. Association for Computational Linguistics.

\bibitem[{Angeli et~al.(2015)Angeli, Johnson~Premkumar, and Manning}]{angeli-etal-2015-leveraging}
Gabor Angeli, Melvin~Jose Johnson~Premkumar, and Christopher~D. Manning. 2015.
\newblock \href {https://doi.org/10.3115/v1/P15-1034} {Leveraging linguistic structure for open domain information extraction}.
\newblock In \emph{Proceedings of the 53rd Annual Meeting of the Association for Computational Linguistics and the 7th International Joint Conference on Natural Language Processing (Volume 1: Long Papers)}, pages 344--354, Beijing, China. Association for Computational Linguistics.

\bibitem[{Bagga and Baldwin(1998)}]{Bagga98algorithmsfor}
Amit Bagga and Breck Baldwin. 1998.
\newblock Algorithms for scoring coreference chains.
\newblock In \emph{In The First International Conference on Language Resources and Evaluation Workshop on Linguistics Coreference}, pages 563--566.

\bibitem[{Barhom et~al.(2019)Barhom, hwartz, Eirew, Bugert, Reimers, and Dagan}]{barhom-etal-2019-revisiting}
Shany Barhom, Vered hwartz, Alon Eirew, Michael Bugert, Nils Reimers, and Ido Dagan. 2019.
\newblock Revisiting joint modeling of cross-document entity and event coreference resolution.
\newblock In \emph{Proceedings of the 57th Annual Meeting of the Association for Computational Linguistics}, pages 4179--4189, Florence, Italy. Association for Computational Linguistics.

\bibitem[{Barry et~al.(2025)Barry, Caillaut, Halftermeyer, Qader, Mouayad, Le~Deit, Cariolaro, and Gesnouin}]{barry-etal-2025-graphrag}
Mariam Barry, Gaetan Caillaut, Pierre Halftermeyer, Raheel Qader, Mehdi Mouayad, Fabrice Le~Deit, Dimitri Cariolaro, and Joseph Gesnouin. 2025.
\newblock \href {https://aclanthology.org/2025.genaik-1.6/} {{G}raph{RAG}: Leveraging graph-based efficiency to minimize hallucinations in {LLM}-driven {RAG} for finance data}.
\newblock In \emph{Proceedings of the Workshop on Generative AI and Knowledge Graphs (GenAIK)}, pages 54--65, Abu Dhabi, UAE. International Committee on Computational Linguistics.

\bibitem[{Bowman et~al.(2015)Bowman, Angeli, Potts, and Manning}]{bowman-etal-2015-large}
Samuel~R. Bowman, Gabor Angeli, Christopher Potts, and Christopher~D. Manning. 2015.
\newblock \href {https://doi.org/10.18653/v1/D15-1075} {A large annotated corpus for learning natural language inference}.
\newblock In \emph{Proceedings of the 2015 Conference on Empirical Methods in Natural Language Processing}, pages 632--642, Lisbon, Portugal. Association for Computational Linguistics.

\bibitem[{Bugert and Gurevych(2021)}]{bugert-gurevych-2021-event}
Michael Bugert and Iryna Gurevych. 2021.
\newblock \href {https://doi.org/10.18653/v1/2021.emnlp-main.38} {{E}vent coreference data (almost) for free: {M}ining hyperlinks from online news}.
\newblock In \emph{Proceedings of the 2021 Conference on Empirical Methods in Natural Language Processing}, pages 471--491, Online and Punta Cana, Dominican Republic. Association for Computational Linguistics.

\bibitem[{Caciularu et~al.(2021)Caciularu, Cohan, Beltagy, Peters, Cattan, and Dagan}]{caciularu-etal-2021-cdlm-cross}
Avi Caciularu, Arman Cohan, Iz~Beltagy, Matthew Peters, Arie Cattan, and Ido Dagan. 2021.
\newblock \href {https://doi.org/10.18653/v1/2021.findings-emnlp.225} {{CDLM}: Cross-document language modeling}.
\newblock In \emph{Findings of the Association for Computational Linguistics: EMNLP 2021}, pages 2648--2662, Punta Cana, Dominican Republic. Association for Computational Linguistics.

\bibitem[{Cattan et~al.(2021{\natexlab{a}})Cattan, Eirew, Stanovsky, Joshi, and Dagan}]{cattan-etal-2021-cross-document}
Arie Cattan, Alon Eirew, Gabriel Stanovsky, Mandar Joshi, and Ido Dagan. 2021{\natexlab{a}}.
\newblock \href {https://doi.org/10.18653/v1/2021.findings-acl.453} {Cross-document coreference resolution over predicted mentions}.
\newblock In \emph{Findings of the Association for Computational Linguistics: ACL-IJCNLP 2021}, pages 5100--5107, Online. Association for Computational Linguistics.

\bibitem[{Cattan et~al.(2021{\natexlab{b}})Cattan, Eirew, Stanovsky, Joshi, and Dagan}]{cattan-etal-2021-realistic}
Arie Cattan, Alon Eirew, Gabriel Stanovsky, Mandar Joshi, and Ido Dagan. 2021{\natexlab{b}}.
\newblock \href {https://doi.org/10.18653/v1/2021.starsem-1.13} {Realistic evaluation principles for cross-document coreference resolution}.
\newblock In \emph{Proceedings of *SEM 2021: The Tenth Joint Conference on Lexical and Computational Semantics}, pages 143--151, Online. Association for Computational Linguistics.

\bibitem[{Chan et~al.(2020)Chan, Schweter, and M{\"{o}}ller}]{chan-etal-2020-germans}
Branden Chan, Stefan Schweter, and Timo M{\"{o}}ller. 2020.
\newblock \href {http://arxiv.org/abs/2010.10906} {German's next language model}.
\newblock \emph{CoRR}, abs/2010.10906.

\bibitem[{Chen et~al.(2025)Chen, Li, and Zhu}]{chen-etal-2025-employing}
Xinyu Chen, Peifeng Li, and Qiaoming Zhu. 2025.
\newblock \href {https://doi.org/10.18653/v1/2025.acl-long.1134} {Employing discourse coherence enhancement to improve cross-document event and entity coreference resolution}.
\newblock In \emph{Proceedings of the 63rd Annual Meeting of the Association for Computational Linguistics (Volume 1: Long Papers)}, pages 23272--23286, Vienna, Austria. Association for Computational Linguistics.

\bibitem[{Chua(2009)}]{Chua2009}
Alton~Y.K. Chua. 2009.
\newblock \href {https://doi.org/10.1108/13673270910971806} {The dark side of successful knowledge management initiatives}.
\newblock \emph{Journal of Knowledge Management}, 13(4):32--40.

\bibitem[{Cybulska and Vossen(2014)}]{cybulska-vossen-2014-using}
Agata Cybulska and Piek Vossen. 2014.
\newblock \href {https://aclanthology.org/L14-1646/} {Using a sledgehammer to crack a nut? lexical diversity and event coreference resolution}.
\newblock In \emph{Proceedings of the Ninth International Conference on Language Resources and Evaluation ({LREC}`14)}, pages 4545--4552, Reykjavik, Iceland. European Language Resources Association (ELRA).

\bibitem[{Devlin et~al.(2019)Devlin, Chang, Lee, and Toutanova}]{devlin-etal-2019-bert}
Jacob Devlin, Ming-Wei Chang, Kenton Lee, and Kristina Toutanova. 2019.
\newblock {BERT}: Pre-training of deep bidirectional transformers for language understanding.
\newblock In \emph{Proceedings of the 2019 Conference of the North {A}merican Chapter of the Association for Computational Linguistics: Human Language Technologies, Volume 1 (Long and Short Papers)}, pages 4171--4186, Minneapolis, Minnesota. Association for Computational Linguistics.

\bibitem[{Eirew et~al.(2021)Eirew, Cattan, and Dagan}]{eirew-etal-2021-wec}
Alon Eirew, Arie Cattan, and Ido Dagan. 2021.
\newblock \href {https://doi.org/10.18653/v1/2021.naacl-main.198} {{WEC}: Deriving a large-scale cross-document event coreference dataset from {W}ikipedia}.
\newblock In \emph{Proceedings of the 2021 Conference of the North American Chapter of the Association for Computational Linguistics: Human Language Technologies}, pages 2498--2510, Online. Association for Computational Linguistics.

\bibitem[{Gao et~al.(2024)Gao, Li, Meng, Li, Zhou, Li, Teng, and Ji}]{gao-etal-2024-enhancing}
Qiang Gao, Bobo Li, Zixiang Meng, Yunlong Li, Jun Zhou, Fei Li, Chong Teng, and Donghong Ji. 2024.
\newblock \href {https://aclanthology.org/2024.lrec-main.523/} {Enhancing cross-document event coreference resolution by discourse structure and semantic information}.
\newblock In \emph{Proceedings of the 2024 Joint International Conference on Computational Linguistics, Language Resources and Evaluation (LREC-COLING 2024)}, pages 5907--5921, Torino, Italia. ELRA and ICCL.

\bibitem[{Huang et~al.(2000)}]{huang2000anaphora}
Yan Huang et~al. 2000.
\newblock \emph{Anaphora: A cross-linguistic approach}.
\newblock Oxford University Press on Demand.

\bibitem[{Lee et~al.(2017)Lee, He, Lewis, and Zettlemoyer}]{lee-etal-2017-end}
Kenton Lee, Luheng He, Mike Lewis, and Luke Zettlemoyer. 2017.
\newblock End-to-end neural coreference resolution.
\newblock In \emph{Proceedings of the 2017 Conference on Empirical Methods in Natural Language Processing}, pages 188--197, Copenhagen, Denmark. Association for Computational Linguistics.

\bibitem[{Lewis et~al.(2020)Lewis, Perez, Piktus, Petroni, Karpukhin, Goyal, K\"{u}ttler, Lewis, Yih, Rockt\"{a}schel, Riedel, and Kiela}]{Lewis2020}
Patrick Lewis, Ethan Perez, Aleksandra Piktus, Fabio Petroni, Vladimir Karpukhin, Naman Goyal, Heinrich K\"{u}ttler, Mike Lewis, Wen-tau Yih, Tim Rockt\"{a}schel, Sebastian Riedel, and Douwe Kiela. 2020.
\newblock \href {https://proceedings.neurips.cc/paper_files/paper/2020/file/6b493230205f780e1bc26945df7481e5-Paper.pdf} {Retrieval-augmented generation for knowledge-intensive nlp tasks}.
\newblock In \emph{Advances in Neural Information Processing Systems}, volume~33, pages 9459--9474. Curran Associates, Inc.

\bibitem[{Luo(2005)}]{luo-2005-coreference}
Xiaoqiang Luo. 2005.
\newblock On coreference resolution performance metrics.
\newblock In \emph{Proceedings of Human Language Technology Conference and Conference on Empirical Methods in Natural Language Processing}, pages 25--32, Vancouver, British Columbia, Canada. Association for Computational Linguistics.

\bibitem[{Mayfield et~al.(2009)Mayfield, Alexander, Dorr, Eisner, Elsayed, Finin, Fink, Freedman, Garera, McNamee, and Mohammad}]{mayfield2009cross}
John Mayfield, Daniel Alexander, Bonnie~J Dorr, Jason Eisner, Tarek Elsayed, Tim Finin, Christine Fink, Marc Freedman, Nikesh Garera, Paul McNamee, and Saif~M Mohammad. 2009.
\newblock Cross-document coreference resolution: A key technology for learning by reading.
\newblock In \emph{AAAI Spring Symposium: Learning by Reading and Learning to Read}, volume~9, pages 65--70.

\bibitem[{Nath et~al.(2024)Nath, Jamil, Ahmed, Baker, Ghosh, Martin, Blanchard, and Krishnaswamy}]{nath-etal-2024-multimodal}
Abhijnan Nath, Huma Jamil, Shafiuddin~Rehan Ahmed, George~Arthur Baker, Rahul Ghosh, James~H. Martin, Nathaniel Blanchard, and Nikhil Krishnaswamy. 2024.
\newblock \href {https://aclanthology.org/2024.lrec-main.1039/} {Multimodal cross-document event coreference resolution using linear semantic transfer and mixed-modality ensembles}.
\newblock In \emph{Proceedings of the 2024 Joint International Conference on Computational Linguistics, Language Resources and Evaluation (LREC-COLING 2024)}, pages 11901--11916, Torino, Italia. ELRA and ICCL.

\bibitem[{Pradhan et~al.(2012)Pradhan, Moschitti, Xue, Uryupina, and Zhang}]{pradhan-etal-2012-conll}
Sameer Pradhan, Alessandro Moschitti, Nianwen Xue, Olga Uryupina, and Yuchen Zhang. 2012.
\newblock {C}o{NLL}-2012 shared task: Modeling multilingual unrestricted coreference in {O}nto{N}otes.
\newblock In \emph{Joint Conference on {EMNLP} and {C}o{NLL} - Shared Task}, pages 1--40, Jeju Island, Korea. Association for Computational Linguistics.

\bibitem[{Recasens and Hovy(2011)}]{10.1017/S135132491000029X}
M.~Recasens and E.~Hovy. 2011.
\newblock Blanc: Implementing the rand index for coreference evaluation.
\newblock \emph{Nat. Lang. Eng.}, 17(4):485–510.

\bibitem[{Reimers and Gurevych(2019)}]{reimers-gurevych-2019-sentence}
Nils Reimers and Iryna Gurevych. 2019.
\newblock \href {https://doi.org/10.18653/v1/D19-1410} {Sentence-{BERT}: Sentence embeddings using {S}iamese {BERT}-networks}.
\newblock In \emph{Proceedings of the 2019 Conference on Empirical Methods in Natural Language Processing and the 9th International Joint Conference on Natural Language Processing (EMNLP-IJCNLP)}, pages 3982--3992, Hong Kong, China. Association for Computational Linguistics.

\bibitem[{Vilain et~al.(1995)Vilain, Burger, Aberdeen, Connolly, and Hirschman}]{10.3115/1072399.1072405}
Marc Vilain, John Burger, John Aberdeen, Dennis Connolly, and Lynette Hirschman. 1995.
\newblock A model-theoretic coreference scoring scheme.
\newblock In \emph{Proceedings of the 6th Conference on Message Understanding}, MUC6 '95, page 45–52, USA. Association for Computational Linguistics.

\bibitem[{Yu et~al.(2022)Yu, Yin, and Roth}]{yu-etal-2022-pairwise}
Xiaodong Yu, Wenpeng Yin, and Dan Roth. 2022.
\newblock \href {https://doi.org/10.18653/v1/2022.starsem-1.6} {Pairwise representation learning for event coreference}.
\newblock In \emph{Proceedings of the 11th Joint Conference on Lexical and Computational Semantics}, pages 69--78, Seattle, Washington. Association for Computational Linguistics.

\bibitem[{Zeng et~al.(2020)Zeng, Jin, Guan, Guo, and Cheng}]{zeng-etal-2020-event}
Yutao Zeng, Xiaolong Jin, Saiping Guan, Jiafeng Guo, and Xueqi Cheng. 2020.
\newblock Event coreference resolution with their paraphrases and argument-aware embeddings.
\newblock In \emph{Proceedings of the 28th International Conference on Computational Linguistics}, pages 3084--3094, Barcelona, Spain (Online). International Committee on Computational Linguistics.

\bibitem[{Zhang et~al.(2024)Zhang, Zhang, Long, Xie, Dai, Tang, Lin, Yang, Xie, Huang, Zhang, Li, and Zhang}]{zhang-etal-2024-mgte}
Xin Zhang, Yanzhao Zhang, Dingkun Long, Wen Xie, Ziqi Dai, Jialong Tang, Huan Lin, Baosong Yang, Pengjun Xie, Fei Huang, Meishan Zhang, Wenjie Li, and Min Zhang. 2024.
\newblock \href {https://doi.org/10.18653/v1/2024.emnlp-industry.103} {{mGTE}: Generalized long-context text representation and reranking models for multilingual text retrieval}.
\newblock In \emph{Proceedings of the 2024 Conference on Empirical Methods in Natural Language Processing: Industry Track}, pages 1393--1412, Miami, Florida, US. Association for Computational Linguistics.

\bibitem[{Zhukova et~al.(2021)Zhukova, Hamborg, Donnay, and Gipp}]{Zhukova2021}
Anastasia Zhukova, Felix Hamborg, Karsten Donnay, and Bela Gipp. 2021.
\newblock Concept identification of directly and indirectly related mentions referring to groups of persons.
\newblock In \emph{Diversity, Divergence, Dialogue}, pages 514--526, Cham. Springer International Publishing.

\bibitem[{Zhukova et~al.(2025{\natexlab{a}})Zhukova, Matt, and Gipp}]{zhukova-etal-2025-automated}
Anastasia Zhukova, Christian~E. Matt, and Bela Gipp. 2025{\natexlab{a}}.
\newblock \href {https://aclanthology.org/2025.loreslm-1.8/} {Automated collection of evaluation dataset for semantic search in low-resource domain language}.
\newblock In \emph{Proceedings of the First Workshop on Language Models for Low-Resource Languages}, pages 112--122, Abu Dhabi, United Arab Emirates. Association for Computational Linguistics.

\bibitem[{Zhukova et~al.(2025{\natexlab{b}})Zhukova, Matt, and Gipp}]{zhukova-2025-efficient}
Anastasia Zhukova, Christian~E. Matt, and Bela Gipp. 2025{\natexlab{b}}.
\newblock Efficient domain-adaptive continual pretraining for the process industry in the german language.
\newblock In \emph{Text, Speech and Dialogue. Proceedings of the 28th International Conference TSD2025, Erlangen, Germany, August 2025}, Cham. Springer Nature Switzerland.

\bibitem[{Zhukova et~al.(2024)Zhukova, von Sperl, Matt, and Gipp}]{Zhukova2024}
Anastasia Zhukova, Lukas von Sperl, Christian~E. Matt, and Bela Gipp. 2024.
\newblock \href {https://doi.org/10.1007/s10115-024-02212-5} {Generative user-experience research for developing domain-specific natural language processing applications}.
\newblock \emph{Knowledge and Information Systems}, 66:7859–7889.

\end{thebibliography}


\end{document}